\documentclass[10pt]{iopart}

\usepackage[table]{xcolor}
\definecolor{Gr}{gray}{0.9}
\usepackage{bm}
\usepackage{graphicx,stackengine}
\usepackage[percent]{overpic}
\usepackage{hyperref}
\hypersetup{
    colorlinks=true,
    linkcolor=blue,
    filecolor=magenta,      
    urlcolor=cyan,
    }
\usepackage{xcolor}

\usepackage{multirow}

\usepackage{tikz}
\usepackage{tikz-qtree}
\usetikzlibrary{trees}	
\usepackage{amsmath} 
\begin{document}
\title[The stochastic human]{The stochastic digital human is now enrolling for in silico imaging trials -- Methods and tools for generating digital cohorts}
\author{A Badano$^1$, M Lago$^1$, E Sizikova$^1$, JG Delfino$^1$, S Guan$^1$ and MA Anastasio$^2$ and B Sahiner$^1$}

\address{$^1$Division of Imaging, Diagnostics, and Software Reliability, Office of Science and Engineering Laboratories, Center for Devices and Radiological Health, \\U. S. Food and Drug Administration, Silver Spring, MD 20993}
\address{$^2$Department of Bioengineering, The Grainger College of Engineering, University of Illinois, Urbana, IL 61801}
\ead{aldo.badano@fda.hhs.gov}
\vspace{10pt}
\begin{indented}
\item[]\today
\end{indented}

\begin{abstract}
 Randomized clinical trials, while often viewed as the highest evidentiary bar by which to judge the quality of a medical intervention, are far from perfect. In silico imaging trials are computational studies that seek to ascertain the performance of a medical device by collecting this information entirely via computer simulations. The benefits of in silico trials for evaluating new technology include significant resource and time savings, minimization of subject risk, the ability to study devices that are not achievable in the physical world, allow for the rapid and effective investigation of new technologies and ensure representation from all relevant subgroups. To conduct in silico trials, digital representations of humans are needed. We review the latest developments in methods and tools for obtaining digital humans for in silico imaging studies. First, we introduce terminology and a classification of digital human models. Second, we survey available methodologies for generating digital humans with healthy and diseased status, and examine briefly the role of augmentation methods. Finally, we discuss the trade-offs of four approaches for sampling digital cohorts and the associated potential for study bias with selecting specific patient distributions.

 \emph{
 Social media blur (100-w): 
 From digital twins to other digital humans for in silico trials: we review methods and tools for obtaining stochastic humans for digital cohorts [LINK]} 
\end{abstract}

\submitto{PRGB}
\maketitle

\tableofcontents
\ioptwocol

\newpage
\section{Introduction}
Two decades ago, in the epilogue of their seminal textbook on image science~\cite{barrett2013foundations}, Barrett and Myers pointed out that in the future, sport games might be played with simulated athletes. The advancement of computer graphics and simulation technologies sparked the notion that perhaps the excitement of a real-life sports event could be conducted in the simulation space with digital models of athletes. Since then, continuous advances in computer processing power and modeling techniques have taken place, driven primarily by entertainment applications~\cite{magnenat2005handbook} and quickly becoming a significant component of research and development (R\&D) efforts in a variety of industries\footnote{To date, Superbowl games are played with physical-world athletes, in part due to the difficulty of conveying real-life personal struggle, an essential component of the entertainment context for sport players and teams (see, for instance, \href{https://www.si.com/soccer/liverpool/articles/darwins-nunezs-incredible-and-inspirational-journey-to-liverpool}{\nolinkurl{here}}).}. Industries that have widely adopted computational modeling and in silico methods throughout the product life-cycle include automotive~\cite{dosovitskiy2017carla} and manufacturing~\cite{cimino2019review} among others~\cite{tao2018digital}. Medicine lags considerably behind~\cite{DT} due, in part, to model complexity, challenging validation, associated potential risks for new devices and drugs, and lack of consensus and regulatory standards. 

Randomized clinical trials, while often viewed as the highest evidentiary bar by which to judge the quality of a medical intervention, are far from perfect. Common causes of failure include safety issues, difficulties with patient recruitment, enrollment, and retention~\cite{fan2022quality}. In addition, clinical trials can suffer from under-representation of rare subpopulations~\cite{us2022diversity}. These limitations represent a unique opportunity to develop in silico trials that are completed as planned, safely, and that include digital cohorts with a representative distribution of subject characteristics and numbers large enough for appropriate statistical power. As pointed out in \cite{rajotte2022synthetic}, in silico data has the potential to address lack of data availability, sharing mechanisms and privacy challenges associated with the use of medical information. 

In silico imaging trials are computational studies that seek to ascertain the performance of a medical device for the intended population, collecting this information entirely in the digital world via computer simulations. The benefits of in silico imaging trials for evaluating new technology include significant resource and time savings, minimization of subject risk, and ethical considerations~\cite{abadi2020virtual,badano2021silico}. Moreover, in silico trials can be used to study devices that do not yet exist or are not practically attainable in the (limited) physical world, allow for the rapid and effective investigation of new technologies~\cite{badano2021silico,Abadi2018-je,
Wedlund}, and facilitate representation from all relevant subpopulations. Each one of these benefits is an essential consideration within the context of the regulatory evaluation  of medical technology~\cite{badano2021silico}. 

The realization that computational models of humans would take center stage in medical imaging system assessment is not new. Full optimization of imaging systems for specific medical tasks requires objects (physical or digital) that represent the variability seen in patients. For many decades, scientists have relied on practical and simpler versions of patients~\cite{segars2009mcat}. However, recent advances in computer processing power and simulation methods are now facilitating the development of more detailed and realistic patient models that are based on digital stochastic descriptions of the model components. For instance, a recent report demonstrated the feasibility of an in silico trial, the Virtual Imaging Clinical Trial for Regulatory Evaluation (VICTRE), as an alternative approach to establish regulatory evidence in support of medical imaging products~\cite{Badano2018EvaluationofDigitalBreastTomosynthesisasReplacementofFull-FieldDigitalMammographyUsinganinSilicoImagingTrial}.

There are numerous parallels between digital- and physical-world  trials. Fundamentally, in silico trials must include the same essential elements of well-designed physical-world clinical trials. Firstly, the population of subjects for whom the new device or technology is intended must be defined. The study design must contain clear rules for selection and rejection of subjects from a distribution of healthy and diseased subjects. However, in silico trials are not subject to effects from covariates in patient selection. For instance, a common problem in evaluating screening tests meant for asymptomatic subjects is that a portion of the enrolled population might be symptomatic~\cite{pepe2005evaluating} with the potential for verification bias~\cite{Arifin}. Secondly, when there are two technologies that are being compared, i.e., a new, yet unproven technology and a comparator technology currently in clinical use, both must  be unambiguously defined. A good choice for comparator technology should be associated with accurate representations of the device characteristics as supported by validation studies~\cite{berti2021validate}. Thirdly, the study requires a definition of the users of the device's outcome (i.e., images in the case of an imaging device trial). These first three components reflect the physical intended use of the device under investigation, i.e., the intended populations of subjects, the intended device comparison, and the intended image interpreters that will be using the device in the physical world. Finally, whether physical or digital, the trial design must provide a definition of the primary outcome to be evaluated, a protocol and statistical analysis associated with the trial, and an analysis of the risk and benefits introduced by the device under investigation.

Both physical and in silico studies require enrollment of representative subjects.  In this review, we survey the latest developments in methods and tools for generating the cohorts of digital humans for imaging studies that represent the variability of physical-world subject populations. We refer to the digital cohorts consisting of digital humans (realizations of the digital human models) as ``stochastic humans''. Assessment of new technology and the regulatory evaluation of that technology requires establishing performance levels for intended populations and, therefore, necessitates computational models that allow sampling of the parameter space defining the subject population in the physical world. We propose to name these models digital humans as opposed to digital replicas or twins to avoid confusion.

The review is organized as follows. First, we introduce terminology and representation models regarding the different types of digital humans described throughout the article. Second, we survey available methodologies for generating digital humans with healthy status and for generating diseased cases. Then, we briefly discuss the role of augmentation methods and conclude with an analysis of sampling techniques that may be used to generate the digital cohorts for evaluating the performance of imaging devices. 

\section{Terminology} A variety of terminologies are being used or proposed for describing digital representations of humans in medicine and other fields. In the literature, some of these are often used without the benefit of a clear definition and, in some instances, wrongly interchangeably. 

We propose to use the term stochastic digital human to denote digital representations of humans (or human body parts) generated from multiple random outputs by sampling known distributions for the model characteristics matching the variability observed in human populations. In contrast, non-stochastic representations are deterministic digital versions of a single physical exemplar (e.g., a model of a human body at a given time) or a group (or family) of physical exemplars which are differentiated by varying physical parameters. Contrary to other terms and concepts currently being discussed including digital families, avatars, chimeras, and digital twins, the concept of a stochastic digital human represents an approach for in silico trials and regulatory evaluation that estimates the performance of an imaging device for a population of subjects rather than for an individual patient, thus incorporating the variability observed in the population.

We propose to classify all digital humans as either individual or population models (see Figure~\ref{class}). Individual models are necessarily image-based while population models can be derived either from images or from knowledge of the fundamental characteristics that define the relevant features of a human.  Note that we will use the term digital human to refer to the models even if the represented object is a part of the body or the whole body of a subject. 

\section{Representations}
Physical objects (including humans) can be represented using continuous variables. We consider the models of humans as continuous in space ($\boldsymbol{r}$) and time (${t}$) and described by a coefficient vector affecting a set of model characteristics: 
\begin{equation} \label{eq1}
    \mathbf{f}_m(\boldsymbol{r},t) \approx \sum_{n=1}^N \theta_n \phi_n(\boldsymbol{r},t).
\end{equation}
Here, $N$ is the dimension of the approximate finite-dimensional representation of the object, and the subscript $m$ indicates the modeling approximation to differentiate from the actual object $\mathbf{f}(\boldsymbol{r},t)$. 

The collection of expansion functions $\{\phi_n(\boldsymbol{r},t)\}_{n=1}^N$ is employed to form $\mathbf{f}_m(\boldsymbol{r},t)$, and $\theta_n$ denotes the $n$-th component of the $N$-dimensional expansion coefficient vector $\boldsymbol{\theta}$. The quantity $\mathbf{f}_m(\boldsymbol{r},t)$ constitutes a discrete representation of a digital human that can be readily displayed on a computer or digitally processed. For the case where the expansion functions  are defined as indicator functions that describe non-overlapping space-time voxels, $\boldsymbol{\theta}$ can sometimes be interpreted as a  digital image whose components $\theta_n$ represent the integrated value of the object over the support of the voxel.

More generally, a  digital human model can be established  by integrating the continuous representation $\mathbf{f}_m(\boldsymbol{r},t)$ over a collection of $N$ voxels as 

\begin{equation}\label{eq2}
    {f}_n = \int_{v_n} \mathbf{f}_m(\boldsymbol{r},{t})\, d^3 \boldsymbol{r}\: d{t}, \hskip 1 cm n=1,\cdots,N,
    \end{equation}
where $v_n$ denotes the support of the $n$-th spatial-temporal voxel and $f_n$ denotes the $n$-th component of a $N$-dimensional vector $\mathbf{f}$ that represents the digital human.

As discussed below, the choice of the expansion functions and associated expansion coefficients can be specified in different ways, with the general goal of making $\mathbf{f}_m(\boldsymbol{r},t)$ an accurate approximation of $\mathbf{f}(\boldsymbol{r},t)$. The expansion functions can depict geometry (e.g., size, morphology), material properties (e.g., x-ray interaction cross-sections, elasticity) or other relevant features (e.g., radioactivity, blood oxygenation levels). For simplicity, we will consider that the stochastic human does not vary with time and proceed only with the spatial dimension $\boldsymbol{r}$. However, the concepts that follow can readily be generalized to model time-varying descriptions.\cite{Barrett2020}

    
\label{sec:repS}
In practice, the coefficient vector $\boldsymbol{\theta}$ can be modeled as a random vector and the expansion functions $\{\phi_n(\boldsymbol{r})\}_{n=1}^N$ as random processes. Methodologies for generating large cohorts of digital stochastic models of humans for in silico imaging trials, including models for organs and tissues with appropriate variability, can rely on either sampling $\boldsymbol{\theta}$, $\phi_n$ or both from appropriate distributions representing the intended population. We can denote the cohort of digital stochastic humans as follows, 
\begin{equation}\label{eq3}
\{\mathbf{f}_s\}_{s=1}^S
= \sum_n \theta_n^s \phi_n(\boldsymbol{r}) , 
    \end{equation}
where $s$ denotes a particular state or random realization of a digital human in a cohort of size $S$.

When $\phi_n$ are known, analytically or numerically, the stochastic models are referred to as procedural. In this case, the modeler is left with choosing the coefficient vector defining the object ($\boldsymbol{\theta}$). In cases for which the defining characteristics are unknown, $\theta_n$ and $\phi_n$ can be estimated from imaging data. 

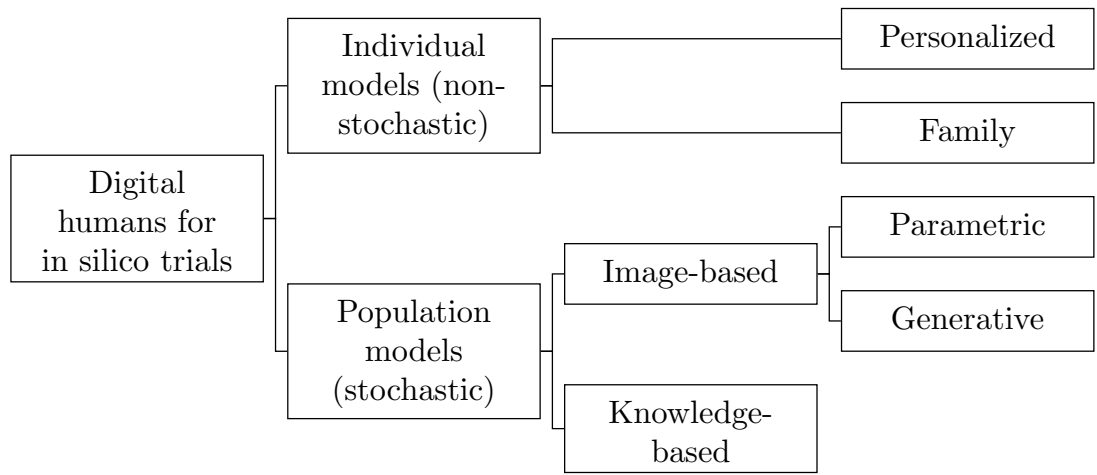
\begin{figure*}
\centering
\resizebox{1.75\columnwidth}{!}{
\begin{tikzpicture}
\tikzset{
  grow'=right,level distance=30mm, sibling distance =3.5mm,
  execute at begin node=\strut,
  every tree node/.style={
			draw,
			anchor = west,
			minimum width=20mm,
			text width=25mm,
			align=center,
			font = {\normalsize}},
         edge from parent/.style={draw, edge from parent fork right}
}
\begin{scope}[frontier/.style={distance from root=90mm}]
\Tree
[.{Digital humans for in silico trials
}
		[.{Individual models (non-stochastic)}
		\node(f1){Personalized};
		\node(f2){Family};
		]
		[.{Population models (stochastic)}
		[.{Image-based}
		    \node(f5){Parametric};
			\node(f6){Generative};
		]
		[.{Knowledge-based}
 		]
 		]
]
\end{scope}
\end{tikzpicture}
}
\caption{Classification of ethods to generate digital humans for in silico clinical trials.}
\label{class}

\end{figure*}

In the following sections, we review available methods and tools for generating digital human models and digital cohorts. We present a classification of available approaches in Figure~\ref{class}.

\section{Individual models}
Individual models attempt to create a digital replica of a specific physical object. Individual models can be categorized as personalized and family models. These models are not stochastic since they are meant to represent individual subjects with as much detail and accuracy as achievable from the image data. In this respect, the representation introduced in Section \ref{sec:repS} applies only with $S=1$ resulting in a single coefficient vector ($\boldsymbol{\theta_n}$) defining the individual.

The digital representation in these cases is typically a multidimensional voxelized array that can be segmented into structures such as tissues and organs. Early attempts relied on geometrical volumes represented by analytical expressions altered to generate a wide variety of sizes and shapes. In other words, $\phi_n$ are described by quadrics and $\theta_n$ represent properties of the volumes defined by the surfaces (e.g., x-ray attenuation and scattering properties). These computational models have proved useful in areas of quality control of imaging systems~\cite{Shepp-Logan74,Martin20} and in radiation dosimetry~\cite{snyder1969estimates}. Even with more sophisticated geometrical structures~\cite{caon2004voxel, zu2005vip, george2014computational} and more spatial detail, these approaches lack the ability to accurately represent the statistical variability found in humans, organs and tissues. While these simpler models remain practical and useful for some tasks, the lack of realism and variability makes them unsuitable for generating digital humans for in silico imaging trials. 

\subsection{Personalized models}
Personalized models aim to capture patient-specific information in a digital representation~\cite{fu2021iphantom}. Medical digital replicas of human subjects are in silico representations of an individual in terms of anatomy and physiology. Sometimes referred to as digital twins~\cite{kamel2021digital}, these replicas are designed to simulate parts or the whole body of a subject for prognostic or predictive assessments. 

These models including digital twins can be continuously updated from
multimodal medical if the data characteristics change over time\footnote{A related concept is an avatar, an artistic and sometimes aspirational digital representation of the human in the digital world for interactivity purposes.}. Digital twins are of interest in the context of evaluating and selecting optimal medical treatments~\cite{Boulos} or imaging procedures~\cite{pesapane2022digital} within clinical practice, and can also be incorporated into other in silico applications~\cite{erol2020digital}. For instance, Wang~\cite{Wang} suggested three applications in the areas of medical imaging: optimal selection of scanning techniques (so called ``virtual comparative scanning''), data sharing from in silico scanning of the digital replica to the open source community, and improvement of the regulatory process of image reconstruction algorithms. Patient image datasets can also be used to generate models of specific tissues and organs. For instance, the Visible Human project~\cite{spitzer1996visible} was first made available in 1994 by the National Library of Medicine (NIH) to facilitate anatomy visualization applications and includes a detailed data set of cross-sectional photographs of the human body. 

\subsection{Family models}
Personalized models of a small number of subjects can be assembled into families to generate a collection of a small number of digital humans spanning a common set of parameters, such as subjects' body size and age. These models are based on image acquisitions using different modalities including computed tomography (CT), magnetic resonance imaging (MRI) and chest radiographs (CXR).

An example of a family model is the Virtual Family~\cite{christ2009virtual}, released by FDA~\footnote{\url{https://www.fda.gov/about-fda/cdrh-offices/virtual-family}} in 2012. The Virtual Family consists of a set of detailed, anatomically correct whole-body models of an adult male, an adult female, and two children based on high-resolution MRI data of healthy volunteers. Organs and tissues are represented using computer-aided design (CAD) techniques where each component is a high-resolution, non self-intersecting mesh. In this case, the models are used for electromagnetic, thermal and acoustic simulations in the safety assessment of active and passive medical implants~\cite{Kyoko}. Safety evaluations do not require full sampling of the intended population and can be performed with a small number of exemplars, provided the exemplars adequately cover the needed parameter space. 

Similar approaches are utilized in efforts to provide models of patient anatomy using patient images as the basis for development of cohorts including using MRI and CT images for modeling lungs~\cite{duetschler2022synthetic} and torso~\cite{segars2009mcat}. More recently, image-derived digital and physical models of the breast have been proposed by Kiarashi~\cite{kiarashi2014} and Bliznakova~\cite{bliznakova2003three}. In this approach, a voxelized breast model is derived from patient images through image segmentation for determining the composition of each voxel~\cite{li2009methodology,erickson2016population,hsu2013generation,elangovan2017design,garcia2020realistic, sarno2021dataset,caballo2022patient}. Patient-derived models are limited to the imaging characteristics of the acquisition system and are also affected by the imperfections of the segmentation methods. The resulting models can also be augmented with physiological features to facilitate imaging studies involving contrast agents~\cite{sauer2022anatomically}.

\section{Population models}

Testing new imaging devices, however, requires the availability of large digital cohorts of stochastic digital humans that can be assembled to properly power a study not only on the aggregate (i.e., for the entire population), but also to analyze for specific subgroups with notable characteristics, including under-represented populations. In this section, we focus our attention on models suited for the generation of large cohorts of digital humans to be enrolled within in silico imaging trials.

\subsection{Image-based models} 
Image-based models estimate and sample model components from relevant characteristics within the acquired medical images. Image-based models estimate model components $\phi_n$ and $\theta_n$ in Eq.~\ref{eq3} from within the acquired medical images.  Whether parametric or generative, all image-based models are limited by the quality of the source data (i.e. medical images), including noise, artifacts, and contrast constraints, and do not provide an unequivocal mapping to the underlying tissues. In practice, the use of image-based models should also acknowledge the limitation arising from the existence of a null space of the imaging system~\cite{tam2012null}. The null space, which typically arises from the mapping of a continuous object to discrete data with an imperfect image acquisition system, results in an unavoidable loss of information regarding the object. Given that the imaging system operator is only partially known for most imaging systems and cannot represent information obscured by the null space of the imaging transformation, image-based models are limited even when imaging system models include noise measurement. 

\subsubsection{Image-based parametric models}\label{sec:image_based_parametric} \quad In image-based parametric models, the generation of cohorts is achieved by creating models based on available sets of patient imaging data and model modification techniques including parametric deformation, morphing, and registration. Parametric models (also known as stylized phantoms~\cite{lee2007hybrid}) capture a population cohort by a set of mathematical equations representing a series of surfaces (e.g., splines) defining organs that are later voxelized into a volumetric model. The popular 4D extended cardiac-torso (XCAT) phantom~\cite{segars20104d} is an example of an image-based parametric model, and a survey of other representations can be found in Kainz~\cite{kainz2018advances}.  

One limitation of this approach is that model development is typically performed on a small number of available patient images. For instance, Erickson~\cite{erickson2016population} presented a methodology to create a database of anatomically variable 3D digital breast models from dedicated breast CT images using a tissue classification and segmentation algorithm and a fuzzy C-means segmentation algorithm. The study provided a population of 224 breast phantoms incorporating a range of breast types, volumes, densities, and parenchymal patterns. However, using hundreds of images might be insufficient to properly characterize a population for statistically powered in silico imaging trials across patient variability. 

Some recently released image datasets include a larger number of cases. For example, the Medical Information Mart for Intensive Care (MIMIC) CXR dataset~\cite{johnson2019mimic} contains 227,835 imaging studies from 65,379 patients presenting to the Beth Israel Deaconess Medical Center Emergency Department between 2011–2016. Similarly, the Medical Imaging and Data Resource Center (MIDRC) effort~\cite{midrc} is undertaking a large, multi-year, systematic effort to collect high-quality COVID data, and over 100,000 imaging studies have been made public after 2 years of work and with significant funding from the NIH. 

However, data sets collected in these well-defined areas are likely still insufficient to capture the total variability in patient images and the large number of subgroups one may find interesting to study~\footnote{``I cannot breed them. So help me, I have tried. We need more \dots than can ever be assembled. Millions, so we can be trillions more,'' Niander Wallace in Blade Runner 2049 (see \url{https://www.imdb.com/title/tt1856101/characters/nm0001467}).}. This limitation precludes the use of image-based parametric models for accurately creating digital cohorts for large scale in silico trials. 

Generation of multiple realizations of humans to constitute a cohort can be obtained by extending image-derived models to create populations in a statistical manner. For instance, Sturgeon~\cite{Sturgeon2017} developed synthetic breast models using principal component analysis (PCA) to describe a small training set of patient images. In this approach, each existing patient breast CT volume was compactly represented by the mean image plus a weighted sum of eigenbreasts. The distribution of weights was sampled to create synthesized breast phantoms that matched fibroglandular density and noise power law exponent distributions in real images. Hence, the distribution of the synthetic model is determined by that of the training data, and, therefore, might suffer from a lack of appropriate representations of cases at the tails of the distribution (e.g., very large or very small, very dense or very glandular breasts). A related concept from the computer vision and graphics community is the statistical human body model, in which a vertex-based model of the body surface is learned, typically via PCA, from subjects' input. The techniques rely on linear blend skinning (LBS) to constrain the surface vertex deformation with respect to a template bone skeleton~\cite{lewis2000pose}. Created for non-medical purposes, these parametric models are typically learned from training examples of lower resolution than what is common in medical imaging. 

One alternative approach is to add deformation morphing using an anatomic template~\cite{fu2021iphantom}. Lee~\cite{lee2007hybrid} introduce a hybrid, non-uniform rational B-spline surface (NURBS) based phantom of an infant by combining the expressiveness of a voxel phantom with the flexibility of geometric manipulation and organ positioning in a parametric phantom. Another example is the XCAT Warp~\cite{chen2020generating}, where AI-assisted unsupervised registration is used to warp XCAT to patient CT images to capture a more broad set of variations, compared to the existing organ and model scaling offered by XCAT. These methods are suitable for investigating digital-twin approaches where individual models reflecting the characteristics of a single individual are needed. 

\subsubsection{Image-based generative models}\quad 
\label{section:gans}
Image-based generative models attempt to synthesize a population of stochastic digital humans from information contained in medical images.  Ideally this population captures the variability in the anatomy and tissue properties within a specified cohort of to-be-imaged subjects. Consider a collection of $N$-dimensional digital humans $\{\mathbf{f}_s\}_{s=1}^S$ that represents the cohort of interest as described by Eq.~\ref{eq3}.
This setting corresponds to a practical situation in which an in silico study employs a fully discrete representation of an imaging system in which a finite-dimensional approximation of an object is mapped to discrete image data.
As mentioned  in Section~\ref{sec:repS}, each digital human $\mathbf{f}_s$ can be interpreted as a realization of a random vector $\mathbf{f}$ that is characterized by an unknown probability density function ${\rm{pr}}(\mathbf{f})$. The ability to sample from ${\rm{pr}}(\mathbf{f})$ to generate large ensembles of objects for use in in silico imaging trials is, at least conceptually, the ultimate objective of a stochastic digital human model. Emerging generative methods that utilize neural networks are being actively developed for this purpose~\cite{galbusera2018exploring}. We refer to these methods as generative models. A generative adversarial network (GAN) is a type of generative model that has recently been very popular for high-resolution image synthesis~\cite{brock2018large}, image translation~\cite{zhu2017unpaired,isola2017image} and a number of generative image applications~\cite{wang2021generative}. Instead of explicitly modelling ${\rm{pr}}(\mathbf{f})$, which is difficult due to the high dimensionality of $\mathbf{f}$, GANs seek to define a stochastic process for drawing samples. As such, GANs are categorized as implicit generative models. Specifically, GANs operate by mapping samples from an analytically tractable, low-dimensional distribution ${\rm{pr}}(\mathbf{z})$ to the sought after samples of the high-dimensional distribution ${\rm{pr}}(\mathbf{f})$. Typically, ${\rm{pr}}(\mathbf{z})$ is specified as an independent and identically distributed (i.i.d.) standard normal distribution, and therefore, samples of the random vector $\mathbf{z}$ can be readily generated. The mapping is usually implemented via a deep neural network referred to as the \emph{generator}. Simultaneously with generator training, a discriminator network is trained to discriminate between the real and generated examples. Therefore, the training process is adversarial and is approximately solving a min-max optimization problem~\cite{goodfellow2020generative}. In this case, a collection of training data (typically images) are utilized to learn how to sample from an empirical distribution that approximates ${\rm{pr}}(\mathbf{f})$. An excellent review of GAN applications for medical image generation can be found in \cite{singh2021medical}. The adversarial training process for GANs is inherently unstable and can result in a phenomenon known as mode collapse, in which the model fails to sample from certain regions of probability space. In addition, the generated samples are often of low resolution. A number of alternative generative models~\cite{bojanowski2017optimizing,ho2020denoising} have been developed to address these challenges in applications to medical imaging~\cite{li2020federated}. For example, generative latent optimization (GLO)~\cite{bojanowski2017optimizing} trains deep convolutional generators by minimizing a simple reconstruction loss, improving on GAN training instabilities. Diffusion models~\cite{ho2020denoising,croitoru2022diffusion} learn a Markov chain of diffusion steps incrementally adding and subtracting noise from data, significantly outperforming GANs in output image quality~\cite{dhariwal2021diffusion}. To date, almost all studies of deep generative models have focused on synthesizing images rather than object representations.


\subsubsection*{Limitations} There are several significant challenges to employing GANs or other types of deep generative models to establish stochastic human models. 
A fundamental and potentially limiting issue is the fact that a collection of objects $\{\mathbf{f}_s\}_{s=1}^S$ is generally not available. Medical images are degraded by the presence of measurement noise and/or reconstruction artifacts which are a limitation of the imaging system and not representative of the true underlying objects. As such, conventional GANs that are directly trained on degraded images will not learn how to sample from the true distribution of objects. In essence, there is a ``chicken and egg problem'' when seeking to establish stochastic human models via deep generative models. There are two possible ways to circumvent this limitation. First, one can utilize high-quality medical images as surrogates of the objects. For example, in certain tomographic imaging modalities and under specific conditions, images of object properties can be reconstructed and accurately approximate the true object properties.  In this case, GANs are trained in the conventional manner, with images representing the training data. If these images are representative of the desired subject cohort, the GAN has the opportunity to accurately capture object variability. Second,  one can modify the GAN training process to incorporate the image degradation process in training. This approach, referred to as an ambient GAN (AmGAN)~\cite{zhou2022learning}, utilizes a generator network that is augmented with a measurement operator. Objects produced by the generator are mapped to degraded image data, which are then compared with experimental images by the discriminator network. This permits establishment of an implicit generative model that describes object randomness to be learned from indirect and noisy measurements of the objects themselves. In a preliminary study, the AmGAN was explored for establishing stochastic object models from imaging measurements for use in optimizing imaging systems~\cite{zhou2022learning}.
 
While promising, the use of deep generative models for in silico clinical trials is nascent and there remain important topics for future investigation. The objective assessment of these technologies is largely lacking, and there is no consensus regarding what statistical information can be reliably learned. Additionally, current models have largely been applied on 2D images and their extension to three-dimensions is an ongoing topic of research.  Finally, as with any data-driven method for establishing stochastic human models, the presence of an imaging system null space will fundamentally limit the ability of GANs to describe certain components of the to-be-imaged objects. The extent to which the null space can be mitigated also remains a topic of ongoing research ~\cite{zhou2022learning}.

\subsection{Knowledge-based models}

Knowledge-based (also known as procedural) models are constructed by sampling a set of $\phi_n$ and $\theta_n$ in Eq.~\ref{eq3} from distributions representing the relevant characteristics of the models. The characteristics of the distributions are often derived from physical or biological measurements. Procedural models allow for an unlimited number of random realizations of the object, leading to the possibility of creating large cohorts of digital humans including the representation of rare cases, and at varying spatial resolution which can properly account for small structures that might be relevant for the specific imaging task being studied. However, they are usually computationally intensive and require a large number of parameters to be defined and estimated based on prior knowledge. Their accuracy and realism depend on the  parameter combinations and they can sometimes generate completely unrealistic outputs.

Knowledge-based, procedural models are common in modeling breast anatomy for imaging studies. Graff~\cite{graff2016new} proposed a detailed model that begins with defining an outside surface using a quadratic hemisphere shell with a skin layer and nipple area overlaid. The shape of the shell is then adjusted for the overall breast volume and surface curvature. Using a Voronoi segmentation approach, the interior is randomly divided into regions of fat or glandular components, with each glandular component containing a ductal network with terminal duct lobular units. The volume is then filled with Cooper's ligaments, chest muscle, and blood vessels. For the VICTRE trial~\cite{Badano2018EvaluationofDigitalBreastTomosynthesisasReplacementofFull-FieldDigitalMammographyUsinganinSilicoImagingTrial}, the breast model was sampled with a 50-$\mu$m voxel size. The implementation is initiated with a set of random seeds and creates random voxelized breast anatomy objects segmented into nine different tissue types. Several different modeling techniques are employed including a non‐isotropic Voronoi segmentation, recursive tree branching algorithms to generate a ductal tree and vascular network, and Perlin‐noise perturbed random spheroids to create fat lobules. 

A similar effort by Bliznakova~\cite{bliznakova2003three} describes a 3D breast software model for x-ray breast imaging simulations based on a breast external shape, ductal lobular system, Cooper’s ligaments and pectoralis muscle. In this approach, a mammographic background texture is added to the tissue regions. Blood vessels, nerves and lymphatics were not modeled explicitly. A similar, more simplistic approach, was developed by Bakic ~\cite{Bakic2016VirtualToolsfortheEvaluationofBreastImaging-State-Of-TheScienceandFutureDirections} based on two ellipsoidal regions of large scale tissue elements: predominantly adipose tissue and predominantly fibro-glandular tissue. Internal tissue structures within these regions are approximated by a distribution of elements including shells, blobs, and a ductal tree. Similar approaches have been reported for full-body models~\cite{lee2007hybrid}.

\section{Modeling disease} \label{section:disease_models}

Disease states can be incorporated into digital cohorts using image-based methods or object-space models of the condition. The analogy between digital human models and disease models can be established if we consider lesions as continuous variables in space ($\boldsymbol{r}$) and time ($\boldsymbol{t}$), described by a coefficient vector affecting a set of lesion model characteristics. For simplicity, we will consider the disease independent (of the underlying anatomy where the disease is located) and additive. This assumption allows us to represent the disease cases as a sum of the stochastic human model and the disease component, an addition that is typically performed in the voxelized object model or directly within the model images.  We recognize this approach is a known simplification, as disease processes often have significant impact on underlying tissues. 

Analogously to the description provided by Eq.~\ref{eq3}, we can generate a set of disease models $\{\mathbf{d}_s\}$ defined by:
\begin{equation}\label{eq4}
\{\mathbf{d}_s\}_{s=1}^S
= \sum_n \lambda_n^s \psi_n(\boldsymbol{r,t}) , 
    \end{equation}
where $\lambda_n^s$ is a disease characteristics coefficient vector described by the function $\psi_n$ over $N$ parameters. Characteristics that define lesions can include geometric functions (e.g., size, morphology), material properties (e.g., x-ray interaction cross-sections, elasticity) or other relevant features (e.g., radioactivity, blood oxygenation levels). 

Methodologies for generating and incorporating disease into cohorts of digital stochastic models  rely on sampling $\lambda_n$ and $\psi_n$ from appropriate distributions representing the intended population. In some cases, disease models are specific to a given anatomical location or physiology corresponding to a digital human exemplar. In other cases, disease models are independent of the digital healthy human and are simply added or inserted multiple times into models of healthy anatomy. In both cases, diseased subjects are denoted by a cohort of digital stochastic humans with added disease components:
\begin{equation}
\{\mathbf{f}_s\}_{s=1}^S
= \sum_n \theta_n^s \phi_n(\boldsymbol{r})
+ \sum_n \lambda_n^s \psi_n(\boldsymbol{r}) , 
\label{eq5}\end{equation}
where $\{\mathbf{f}_s\}_{s=1}^S$ is a cohort of diseased digital humans (for simplicity, and similarly as in the previous section, we choose to omit the time dimension). Similarly to normal models, when $\psi_n$ are unknown, models of disease can be obtained relying on imaging. Alternatively, when $\psi_n$ are known, analytically or numerically, the stochastic disease models are referred to as knowledge-based (also known as procedural). 

\subsection{Image-based models of disease}
Similar to image-based models of the human body, image-based models of disease rely on imaging data for extracting lesion information. Various techniques for 
capturing disease characteristics, particularly for breast lesions, have recently been explored~\cite{dukov2019models, bliznakova2019development}. Image-based neural network models for disease modelling have also been explored. For instance, Kadia~\cite{Kadia} proposed a method to generate synthetic, infection-like patterns in the lung to create large collections of 2D and 3D training examples for deep segmentation models.
While image-based models contain features from actual patient data and thus may look more realistic at first glance, they suffer from limited resolution of the tumor model, largely determined by the imaging acquisition characteristics and limited number of available lesion morphologies, shapes, and sizes. In addition, image-based methods require an institutional review board (IRB) approval for obtaining and utilizing the diseased case data for research and development, which could delay or disadvantage some analysis efforts. 


\subsection{Knowledge-based models of disease}
Knowledge-based models of disease are constructed by sampling a set of known (or assumed known) $\psi_n$ and $\lambda_n$ in Eq.~\ref{eq4} from distributions representing the relevant characteristics of the disease, where distributions are often derived from physical or biological measurements. In contrast to image-based models, knowledge-based models enable the generation of unlimited numbers of lesion shapes with variable resolution. Examples of knowledge-based models include de Sisternes~\cite{de2015computational} spiculated breast cancer mass model and Sengupta~\cite{sengupta2021computational} growing breast mass models. In \cite{sengupta2021computational}, a breast lesion growth method based on biological and physiological phenomena accounting for the stiffness of surrounding anatomical structures was introduced. Breast ligaments were considered as rigid structures with elastic moduli in the range of 8x$10^4$-4x$10^5$ kPa, while fat (elastic modulus varying from 0.5 to 25 kPa) and glandular tissues (elastic modulus varying from 7.5 to 66 kPa) constituting the more elastic regions of the breast. In this approach, tumor cells are less likely to grow through stiffer structures and instead, preferentially proliferate through the more elastic regions of the breast. Depending on the breast local anatomical structures, a range of unique lesion morphologies can be realized, allowing lesions to blend naturally into the anatomical regions. 

A common simplifying assumption is to define the disease model independent from other human model components. For example, in VICTRE~\cite{Badano2018EvaluationofDigitalBreastTomosynthesisasReplacementofFull-FieldDigitalMammographyUsinganinSilicoImagingTrial} and in Sengupta~\cite{Sengupta2021ComputationalMO}, breast cancer mass lesions are added to the normal breast models by replacing voxels in the breast with voxels of the lesion model, without modification to adjacent voxels. This approach, while practical, does not account for the significant effect of the growing tumors on its surrounding tissues, typically visible in x-ray images as architectural distortions suggestive of abnormalities. To consider these effects, Eq.~\ref{eq5} needs to be modified to account for the interaction between normal and disease models. 

\section{Role of augmentation methods} 
\label{sec:Role_of_augmentation_methods}
Augmentation methods start with an already-defined object, image or a set of defined objects, and generate new examples based on properties of inputs, as well as pre-defined or data-driven transformations (in contrast, digital human models start with only an object description, such as that given in Eq. \ref{eq1}). GAN-based models (see Section~\ref{section:gans}) are similar to augmentation methods in that they employ complex transformations derived with the help of training data sets. Augmentation methods typically employ analytically-defined or stochastic operators that do not require the use of neural networks, and can be applied both in the object domain and in the acquired image domain. Techniques in the latter group generate examples that could be obtained through an imaging system applied to an object with an accompanying degradation (e.g., smoothing, noise, reconstruction artifacts).  

Geometric transformations, intensity operations, and spatial filtering are among the most basic types of augmentation methods. Geometric transformations redefine the spatial relationships among voxels or geometrical locations in an object, and include affine (scaling, rotation, translation, reflection and shearing), as well as non-affine transformations, such as non-linear warping and morphing ~\cite{Wolberg1988GeometricTT}. Intensity operations modify intensity values in a grayscale image or channel values (e.g., RGB or CMYK) in a color image. Examples include operations such as a family of gamma corrections, linear contrast adjustments, and remapping voxel values using a pre-defined or pseudo-random remapping curve~\cite{Chlap2021-xi, Hesse2021-zj}. Spatial filtering (using a filter mask) is another possibility for generating a new image or object based on an existing one. Spatial filtering can be linear (in which case it can be implemented by a convolution operation) or non-linear (e.g., median filtering), and can be implemented to smooth or sharpen to emphasize certain features. Finally, all three types of augmentations can be combined using a continuous mapping from the parameter space of transformations to the image or object space~\cite{NEURIPS2021_7230b2b0}.      

Noise injection is an image augmentation method that enhances robustness of machine learning models and belongs to the family of domain randomization (DR) methods~\cite{Noh2017RegularizingDN}. Although noise injection after data acquisition does not generate a new member of a patient population, it can generate a different representation of an object in the image domain, and can be useful for augmenting patient cohorts obtained with in silico modeling. Some earlier and non-medical applications of noise injection in machine learning sought to augment the image data sets without regard to the physics of image acquisition~\cite{Moreno-Barea2018,Bae2018-kv}. Other works used physics-based techniques for noise modeling and addition, improving realism of the noise appearance in the augmented images~\cite{Omigbodun2019,pmlr-v139-fabian21a}. The main benefit of noise injection in the image domain for in silico trials is that it may allow for the rapid generation of different representations of the same object at different noise levels, leading to comparisons that may require less computational power compared to a full implementation of image acquisition physics applied to a digital stochastic object model. Addition of texture to a model in the object domain  has similarities to noise injection in the image domain in that both techniques aim at producing noise-like properties (e.g., using a noise power spectrum in modeling), but are different in that addition of texture in the object domain does not attempt to model the noise from data acquisition~\cite{Abadi2019}.       

Combination of objects or images is another popular augmentation technique. In the object domain, combination of an object model for a normal (non-diseased) patient with a lesion model (as described in Section~\ref{section:disease_models}) can be thought of as an example of this type of augmentation. Generating new members of a patient population based on an eigenspace analysis of existing patient objects, as was done in \cite{Sturgeon2017} and described in Section~\ref{sec:image_based_parametric} is another example of augmentation in the object domain. In the image domain, researchers investigated tools for the extraction of image parts from one clinical image and then their insertion into a new location on the same or different image. Pezeshk~\cite{Pezeshk2015} used an image blending technique based on Poisson image editing to insert pulmonary nodules extracted from one chest CT exam into another. Augmenting a training data set for a machine learning model using this technique can improve the model performance on independent, real test datasets~\cite{Pezeshk2017}. Likewise, Ghanian~\cite{Ghanian_2018} used a similar technique to insert microcalcification clusters extracted from one mammogram into another mammogram, and showed that  experienced observers cannot reliably distinguish between computationally inserted and native clusters. Besides the ability to convince experts, desirable properties for such combination techniques include acceptable noise properties in the combined image, plausible lesion-background combinations (that might require the intervention of an operator during the augmentation process), and a sufficient range of variation in the combined images that can be generated, which are often difficult to satisfy simultaneously. 

The main advantage of data augmentation methods is their practicality. For example, existing images or models both for normal and diseased patients can be manipulated (with relative ease) with geometric transformations leading to expanded patient representations. When implemented in the image domain, augmentation methods are fast, bypassing the stage where a model for the imaging system is applied to the object to yield an image. However, important shortcomings accompany these advantages. Unless deliberate attention is paid, augmentation methods may yield objects or images that are biologically or physically implausible. An extreme example may be an intensity transformation that results in bones with lower Hounsfield units than soft tissue. Although this can be avoided easily by using an intensity transformation that is monotonically increasing, most augmentation methods and transformations need careful planning to avoid such inconsistencies, and it may not be possible to avoid all inconsistencies. The consequences of such implausible images or objects on the results of an in silico imaging trial should be carefully considered. In addition, many augmentation techniques do not result in an independent, new representation from the population, but rather in representations that are highly dependent on the original objects or images used as inputs to the augmentation method. For example, lesion insertion methods described in the previous paragraph do not increase the number of lesions in the augmented data set, but only the lesion-background combinations that are generated. Again, the consequences of this limitation in the range of variation of generated images should be an important consideration in an in silico imaging trial that uses augmentation.

\section{Considerations for sampling digital cohorts}
\begin{figure*}[t]
   \begin{center}
   \includegraphics[trim={0 8cm 0cm 2cm},clip,page=2,width=1.96\columnwidth]{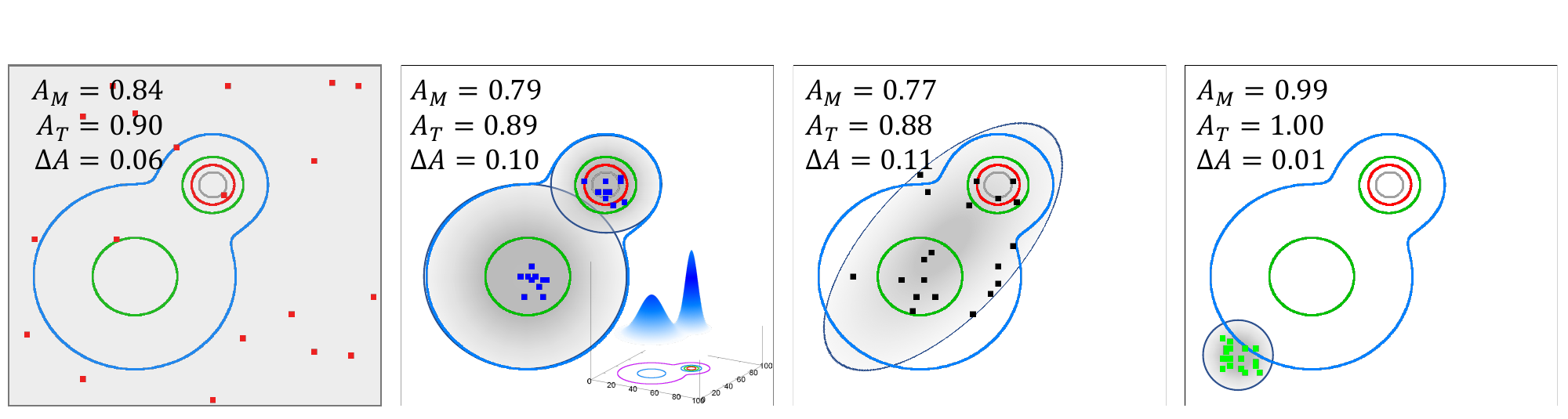}
   \caption{Effect of sampling strategies on performance assessment. Sampling is from a bimodal distribution of subjects (seen in 3D insert in the second panel from the left) described by 2 random parameters: (from left to right) uniform, matched, simpler, and narrow. Only 20 samples are shown here for ease of visualization. The gray shading depicts the distribution from which samples are taken in each of the 4 cases. $A_M$, $A_T$, and $\Delta A$ refer to the lesion detection average AUC for mammography, average AUC for digital breast tomosynthesis, and the average AUC difference, respectively.  \label{fig:sampling}.}
   \end{center}
\end{figure*}

In silico studies require careful study planning and good clinical trial design.  Even if and when methodologies for developing digital stochastic models of humans for imaging studies become widely available, generating digital cohorts needs an understanding of the trade-offs and potential for bias associated with selecting a specific distribution of study subjects. At the start of the design of an in silico imaging trial is the challenging task of scoping the population of the digital humans to be included in the study. For instance, a number of previous computational studies in breast imaging using procedural models used a uniform sampling with a desired average of 50\% adipose and 50\% fibroglandular voxels~\cite{gong2006computer} with an uncompressed breast size of 14~cm. Another example of enrollment strategy can be found in the OpenVCT platform, where a range of size and glandularity is specified and then uniformly randomly sampled~\cite{Barufaldi18}. A more recent in silico imaging study used sampling from a multi-class distribution identifying 4 different breast densities resulting in the characteristics of the intended population~\cite{Badano2018EvaluationofDigitalBreastTomosynthesisasReplacementofFull-FieldDigitalMammographyUsinganinSilicoImagingTrial}.

Through in silico enrollment, digital cohorts $\{\mathbf{f}_s\}_{s=1}^S$ are generated. We denote the distribution of the population of digital humans as 
$\mathbf{f}_d$, where $d$ represents the digital world, and the distribution of subjects in the intended population as $\mathbf{f}_i$. In this context, the goal of the in silico enrollment is to minimize the difference $\Delta\mathbf{f}=|\mathbf{f}_{d}-\mathbf{f}_i|$ between the digital ($d$) and physical-world intended distributions, where $|.|$ denotes a statistical distance measure. Clinical trial enrollment programs in the physical world require strategies to ensure a reasonable $\Delta\mathbf{f}$ given available sampling resources. At first approximation, the in silico enrollment should approximate the intended distribution to a greater extent than the corresponding physical clinical trial enrollment.

Analysis of $\Delta\mathbf{f}$ corresponding to a given in silico enrollment strategy may be needed to understand how the difference across study subject distributions could affect the outcome of the trial. Here, we discuss a test case (see Figure~\ref{fig:sampling}) that compares different enrollment strategies for an in silico trial comparing digital mammography (DM) and digital breast tomosynthesis (DBT) derived from the VICTRE~\cite{Badano2018EvaluationofDigitalBreastTomosynthesisasReplacementofFull-FieldDigitalMammographyUsinganinSilicoImagingTrial} project. We assume the populations (digital and physical) consist of normal and diseased subjects with a prevalence of 0.5. These two classes of patients are therefore sampled with equal probability. We calculate the difference of performance (measured using the area under the receiver operating characteristic curve, or AUC, in the task of differentiating between normal and disease subjects) between mammography and digital breast tomosynthesis. We consider the following four sampling approaches. In the first approach (uniform), $\mathbf{f}_i$ is unknown and subjects are sampled uniformly within a range of interest, from all possible combinations of the input parameters that define $\mathbf{f}$. In the second approach (matched), $\mathbf{f}_i$ is known and subjects are sampled from the true underlying distribution. In the third approach (simpler), $\mathbf{f}_i$ is unknown, but can be approximated by another, simpler distribution from which samples are obtained. Finally, in the fourth approach (narrow), $\mathbf{f}_i$ is known to be a narrow, well-defined subset of the general population of subjects of particular interest (e.g., rare diseases or very obese subjects).

For this simplified example, let $\mathbf{f}_i$ be a bimodal distribution defined by two parameters (e.g., breast size and glandularity). Using Eq.~\ref{eq3}, we can express the model through two expansion functions $\phi_{1,2}$, each associated with one of the two random variables affected by a random parameter set given by $\theta_{1,2}$. As seen in Figure~\ref{fig:sampling}, one of the modes of the distribution has twice the amplitude and half the variance of the other. The four density plots illustrate a top view of the distribution contour plot with the individual samples drawn using the four different sampling strategies. The results demonstrate that the choice of sampling strategy can have a significant effect on the difference in AUC, which for this example case, ranges from a difference of 0.01 (almost zero) to 0.11 in terms of device performance.

\section{Summary and conclusions}
In silico trials are an emerging area of regulatory research that offer the ability to capture highly diverse patient distributions at a significant time and cost savings, compared to traditional physical clinical trials. To conduct in silico trials, realistic digital representations of humans are needed. In this paper, we reviewed and discussed existing techniques for generating digital humans, including disease models, for in silico imaging trials. Digital humans can be created using image-based or knowledge-based techniques.  In summary, we favor techniques with object-based representations (rather than images of objects) in order to decouple the characteristics of the image acquisition system from the characteristics of the object (true representation of the physical-world human). In generating digital humans for in silico trials, one should consider the quality and quantity of the source data or knowledge used, and whether the models represent a single patient, a small cohort, or a sizable population with realistic patient variability. 

It remains a crucial next step to evaluate the quality of the digital human models and the images that can be generated with them. In particular, it is essential to carefully identify the patient distribution that the particular digital human model can and cannot capture, in order to prevent misuse and ensure patient safety. We need to study to what extent model-derived data contributes to our understanding of performance levels for populations with rare diseases or for populations underrepresented in traditional clinical trials. Future work should examine the ethical and safety considerations of relying on digital humans for clinical trials. Overall, the use of in silico imaging trials and in silico trials in medicine is a rapidly developing field and has the potential to address many of the emerging challenges in the regulatory evaluation of medical devices.
\section*{References}
\bibliographystyle{ieeetr}   %
\bibliography{bibliography}

\begin{thebibliography}{10}

\bibitem{barrett2013foundations}
H.~H. Barrett and K.~J. Myers, {\em Foundations of image science}.
\newblock John Wiley \& Sons, 2013.

\bibitem{magnenat2005handbook}
N.~Magnenat-Thalmann and D.~Thalmann, {\em Handbook of virtual humans}.
\newblock John Wiley \& Sons, 2005.

\bibitem{dosovitskiy2017carla}
A.~Dosovitskiy, G.~Ros, F.~Codevilla, A.~Lopez, and V.~Koltun, ``Carla: An open
  urban driving simulator,'' in {\em Conference on robot learning}, pp.~1--16,
  PMLR, 2017.

\bibitem{cimino2019review}
C.~Cimino, E.~Negri, and L.~Fumagalli, ``Review of digital twin applications in
  manufacturing,'' {\em Computers in Industry}, vol.~113, p.~103130, 2019.

\bibitem{tao2018digital}
F.~Tao, H.~Zhang, A.~Liu, and A.~Y. Nee, ``Digital twin in industry:
  State-of-the-art,'' {\em IEEE Transactions on industrial informatics},
  vol.~15, no.~4, pp.~2405--2415, 2018.

\bibitem{DT}
A.~Thelen, X.~Zhang, O.~Fink, and et~al., ``A comprehensive review of digital
  twin - part 1: modeling and twinning enabling technologies,'' {\em Struct
  Multidisc Optim}, 2022.

\bibitem{fan2022quality}
J.~Fan, X.~Liu, Y.~Li, H.~Xia, R.~Yang, J.~Li, and Y.~Zhang, ``Quality problems
  of clinical trials in china: evidence from quality related studies,'' {\em
  Trials}, vol.~23, no.~1, pp.~1--11, 2022.

\bibitem{us2022diversity}
U.~Food, D.~Administration, {\em et~al.}, ``Diversity plans to improve
  enrollment of participants from underrepresented racial and ethnic
  populations in clinical trials; draft guidance for industry; availability,''
  2022.

\bibitem{rajotte2022synthetic}
J.-F. Rajotte, R.~Bergen, D.~L. Buckeridge, K.~El~Emam, R.~Ng, and E.~Strome,
  ``Synthetic data as an enabler for machine learning applications in
  medicine,'' {\em Iscience}, vol.~25, no.~11, 2022.

\bibitem{abadi2020virtual}
E.~Abadi, W.~P. Segars, B.~M. Tsui, P.~E. Kinahan, N.~Bottenus, A.~F. Frangi,
  A.~Maidment, J.~Lo, and E.~Samei, ``Virtual clinical trials in medical
  imaging: a review,'' {\em Journal of Medical Imaging}, vol.~7, no.~4,
  p.~042805, 2020.

\bibitem{badano2021silico}
A.~Badano, ``In silico imaging clinical trials: cheaper, faster, better, safer,
  and more scalable,'' {\em Trials}, vol.~22, no.~1, pp.~1--7, 2021.

\bibitem{Abadi2018-je}
E.~Abadi, W.~P. Segars, G.~M. Sturgeon, J.~E. Roos, C.~E. Ravin, and E.~Samei,
  ``Modeling lung architecture in the {XCAT} series of phantoms:
  Physiologically based airways, arteries and veins,'' {\em IEEE Trans. Med.
  Imaging}, vol.~37, pp.~693--702, Mar. 2018.

\bibitem{Wedlund}
L.~Wedlund and J.~Kvedar, ``Simulated trials: in silico approach adds depth and
  nuance to the rct gold-standard,'' {\em NPJ digital medicine}, vol.~4, no.~1,
  2021.

\bibitem{segars2009mcat}
W.~P. Segars and B.~M. Tsui, ``Mcat to xcat: The evolution of 4-d computerized
  phantoms for imaging research,'' {\em Proceedings of the IEEE}, vol.~97,
  no.~12, pp.~1954--1968, 2009.

\bibitem{Badano2018EvaluationofDigitalBreastTomosynthesisasReplacementofFull-FieldDigitalMammographyUsinganinSilicoImagingTrial}
A.~Badano, C.~G. Graff, A.~Badal, D.~Sharma, R.~Zeng, F.~W. Samuelson, S.~J.
  Glick, and K.~J. Myers, ``Evaluation of digital breast tomosynthesis as
  replacement of full-field digital mammography using an in silico imaging
  trial,'' {\em JAMA network open}, vol.~1, pp.~e185474--e185474, 11 2018.

\bibitem{pepe2005evaluating}
M.~Pepe, ``Evaluating technologies for classification and prediction in
  medicine,'' {\em Statistics in medicine}, vol.~24, no.~24, pp.~3687--3696,
  2005.

\bibitem{Arifin}
W.~N. Arifin and U.~K. Yusof, ``Correcting for partial verification bias in
  diagnostic accuracy studies: A tutorial using r,'' {\em Statistics in
  Medicine}, vol.~41, no.~9, pp.~1709--1727, 2022.

\bibitem{berti2021validate}
F.~Berti, L.~Antonini, G.~Poletti, C.~Fiuza, T.~J. Vaughan, F.~Migliavacca,
  L.~Petrini, and G.~Pennati, ``How to validate in silico deployment of
  coronary stents: strategies and limitations in the choice of comparator,''
  {\em Frontiers in Medical Technology}, p.~37, 2021.

\bibitem{Barrett2020}
H.~H. Barrett and L.~Caucci, ``{Stochastic models for objects and images in
  oncology and virology: application to PI3K-Akt-mTOR signaling and COVID-19
  disease},'' {\em Journal of Medical Imaging}, vol.~8, no.~S1, p.~S16001,
  2020.

\bibitem{Shepp-Logan74}
L.~A. Shepp and B.~F. Logan, ``The fourier reconstruction of a head section,''
  {\em IEEE Transactions on Nuclear Science}, vol.~21, no.~3, pp.~21--43, 1974.

\bibitem{Martin20}
J.~Martin, M.~Ruthven, R.~Boubertakh, and M.~E. Miquel, ``Realistic dynamic
  numerical phantom for mri of the upper vocal tract,'' {\em Journal of
  Imaging}, vol.~6, no.~9, 2020.

\bibitem{snyder1969estimates}
W.~S. Snyder, M.~R. Ford, G.~G. Warner, and H.~Fisher~Jr, ``Estimates of
  absorbed fractions for monoenergetic photon sources uniformly distributed in
  various organs of a heterogeneous phantom.,'' tech. rep., Oak Ridge National
  Lab., Tenn., 1969.

\bibitem{caon2004voxel}
M.~Caon, ``Voxel-based computational models of real human anatomy: a review,''
  {\em Radiation and environmental biophysics}, vol.~42, no.~4, pp.~229--235,
  2004.

\bibitem{zu2005vip}
X.~G. Zu, ``The vip-man model-a digital human testbed for radiation
  siimulations,'' {\em SAE transactions}, pp.~779--787, 2005.

\bibitem{george2014computational}
X.~George~Xu, ``Computational phantoms for organ dose calculations in radiation
  protection and imaging,'' {\em The Phantoms of Medical and Health Physics},
  pp.~225--262, 2014.

\bibitem{fu2021iphantom}
W.~Fu, S.~Sharma, E.~Abadi, A.-S. Iliopoulos, Q.~Wang, J.~Y. Lo, X.~Sun, W.~P.
  Segars, and E.~Samei, ``iphantom: a framework for automated creation of
  individualized computational phantoms and its application to ct organ
  dosimetry,'' {\em IEEE Journal of Biomedical and Health Informatics},
  vol.~25, no.~8, pp.~3061--3072, 2021.

\bibitem{kamel2021digital}
M.~N. Kamel~Boulos and P.~Zhang, ``Digital twins: from personalised medicine to
  precision public health,'' {\em Journal of Personalized Medicine}, vol.~11,
  no.~8, p.~745, 2021.

\bibitem{Boulos}
M.~Kamel~Boulos and Z.~P., ``Digital twins: From personalised medicine to
  precision public health,'' {\em J Pers Med}, vol.~11, no.~8, p.~745, 2021.

\bibitem{pesapane2022digital}
F.~Pesapane, A.~Rotili, S.~Penco, L.~Nicosia, and E.~Cassano, ``Digital twins
  in radiology,'' {\em Journal of Clinical Medicine}, vol.~11, no.~21, p.~6553,
  2022.

\bibitem{erol2020digital}
T.~Erol, A.~F. Mendi, and D.~Do{\u{g}}an, ``The digital twin revolution in
  healthcare,'' in {\em 2020 4th International Symposium on Multidisciplinary
  Studies and Innovative Technologies (ISMSIT)}, pp.~1--7, IEEE, 2020.

\bibitem{Wang}
G.~Wang, A.~Badal, X.~Jia, J.~S. Maltz, J.~Mueller, K.~J. Myers, C.~Niu,
  M.~Vannier, P.~Yan, Z.~Yu, and R.~Zeng, ``Development of metaverse for
  intelligent healthcare,'' {\em Nature Machine Intelligence}, vol.~4,
  p.~9220929, 2022.

\bibitem{spitzer1996visible}
V.~Spitzer, M.~J. Ackerman, A.~L. Scherzinger, and D.~Whitlock, ``The visible
  human male: a technical report,'' {\em Journal of the American Medical
  Informatics Association}, vol.~3, no.~2, pp.~118--130, 1996.

\bibitem{christ2009virtual}
A.~Christ, W.~Kainz, E.~G. Hahn, K.~Honegger, M.~Zefferer, E.~Neufeld,
  W.~Rascher, R.~Janka, W.~Bautz, J.~Chen, {\em et~al.}, ``The virtual family,
  development of surface-based anatomical models of two adults and two children
  for dosimetric simulations,'' {\em Physics in Medicine and Biology}, vol.~55,
  no.~2, p.~N23, 2009.

\bibitem{Kyoko}
K.~Fujimoto, T.~A. Zaidi, D.~Lampman, J.~W. Guag, S.~Etheridge, H.~Habara, and
  S.~S. Rajan, ``Comparison of sar distribution of hip and knee implantable
  devices in 1.5t conventional cylindrical-bore and 1.2t open-bore vertical mri
  systems,'' {\em Magnetic Resonance in Medicine}, vol.~87, no.~3,
  pp.~1515--1528, 2022.

\bibitem{duetschler2022synthetic}
A.~Duetschler, G.~Bauman, O.~Bieri, P.~C. Cattin, S.~Ehrbar, G.~Engin-Deniz,
  A.~Giger, M.~Josipovic, C.~Jud, M.~Krieger, {\em et~al.}, ``Synthetic 4dct
  (mri) lung phantom generation for 4d radiotherapy and image guidance
  investigations,'' {\em Medical physics}, vol.~49, no.~5, pp.~2890--2903,
  2022.

\bibitem{kiarashi2014}
N.~Kiarashi, A.~Nolte, G.~M. Sturgeon, W.~P. Segars, S.~V. Ghate, L.~W. Nolte,
  E.~Samei, and J.~Y. Lo, ``Development and application of a suite of 4-d
  virtual breast phantoms for optimization and evaluation of breast imaging
  systems,'' {\em IEEE Trans Med Imaging}, vol.~33, no.~7, pp.~1401--9, 2014.

\bibitem{bliznakova2003three}
K.~Bliznakova, Z.~Bliznakov, V.~Bravou, Z.~Kolitsi, and N.~Pallikarakis, ``A
  three-dimensional breast software phantom for mammography simulation,'' {\em
  Physics in Medicine \& Biology}, vol.~48, no.~22, p.~3699, 2003.

\bibitem{li2009methodology}
C.~M. Li, W.~P. Segars, G.~D. Tourassi, J.~M. Boone, and J.~T. Dobbins~III,
  ``Methodology for generating a 3d computerized breast phantom from empirical
  data,'' {\em Medical physics}, vol.~36, no.~7, pp.~3122--3131, 2009.

\bibitem{erickson2016population}
D.~W. Erickson, J.~R. Wells, G.~M. Sturgeon, E.~Samei, J.~T. Dobbins, W.~P.
  Segars, and J.~Y. Lo, ``Population of 224 realistic human subject-based
  computational breast phantoms,'' {\em Medical physics}, vol.~43, no.~1,
  pp.~23--32, 2016.

\bibitem{hsu2013generation}
C.~M. Hsu, M.~L. Palmeri, W.~P. Segars, A.~I. Veress, and J.~T. Dobbins~III,
  ``Generation of a suite of 3d computer-generated breast phantoms from a
  limited set of human subject data,'' {\em Medical physics}, vol.~40, no.~4,
  p.~043703, 2013.

\bibitem{elangovan2017design}
P.~Elangovan, A.~Mackenzie, D.~R. Dance, K.~C. Young, V.~Cooke, L.~Wilkinson,
  R.~M. Given-Wilson, M.~G. Wallis, and K.~Wells, ``Design and validation of
  realistic breast models for use in multiple alternative forced choice virtual
  clinical trials,'' {\em Physics in Medicine \& Biology}, vol.~62, no.~7,
  p.~2778, 2017.

\bibitem{garcia2020realistic}
E.~Garc{\'\i}a, C.~Fedon, M.~Caballo, R.~Mart{\'\i}, I.~Sechopoulos, and
  O.~Diaz, ``Realistic compressed breast phantoms for medical physics
  applications,'' in {\em 15th International Workshop on Breast Imaging
  (IWBI2020)}, vol.~11513, pp.~30--37, SPIE, 2020.

\bibitem{sarno2021dataset}
A.~Sarno, G.~Mettivier, F.~di~Franco, A.~Varallo, K.~Bliznakova, A.~M.
  Hernandez, J.~M. Boone, and P.~Russo, ``Dataset of patient-derived digital
  breast phantoms for in silico studies in breast computed tomography, digital
  breast tomosynthesis, and digital mammography,'' {\em Medical Physics},
  vol.~48, no.~5, pp.~2682--2693, 2021.

\bibitem{caballo2022patient}
M.~Caballo, C.~Rabin, C.~Fedon, A.~Rodr{\'\i}guez-Ruiz, O.~Diaz, J.~M. Boone,
  D.~R. Dance, and I.~Sechopoulos, ``Patient-derived heterogeneous breast
  phantoms for advanced dosimetry in mammography and tomosynthesis,'' {\em
  Medical physics}, vol.~49, no.~8, pp.~5423--5438, 2022.

\bibitem{sauer2022anatomically}
T.~J. Sauer, E.~Abadi, P.~Segars, and E.~Samei, ``Anatomically and
  physiologically informed computational model of hepatic contrast perfusion
  for virtual imaging trials,'' {\em Medical Physics}, vol.~49, no.~5,
  pp.~2938--2951, 2022.

\bibitem{tam2012null}
L.~K. Tam, J.~P. Stockmann, G.~Galiana, and R.~T. Constable, ``Null space
  imaging: nonlinear magnetic encoding fields designed complementary to
  receiver coil sensitivities for improved acceleration in parallel imaging,''
  {\em Magnetic resonance in medicine}, vol.~68, no.~4, pp.~1166--1175, 2012.

\bibitem{lee2007hybrid}
C.~Lee, D.~Lodwick, D.~Hasenauer, J.~L. Williams, C.~Lee, and W.~E. Bolch,
  ``Hybrid computational phantoms of the male and female newborn patient:
  Nurbs-based whole-body models,'' {\em Physics in Medicine \& Biology},
  vol.~52, no.~12, p.~3309, 2007.

\bibitem{segars20104d}
W.~P. Segars, G.~Sturgeon, S.~Mendonca, J.~Grimes, and B.~M. Tsui, ``4d xcat
  phantom for multimodality imaging research,'' {\em Medical physics}, vol.~37,
  no.~9, pp.~4902--4915, 2010.

\bibitem{kainz2018advances}
W.~Kainz, E.~Neufeld, W.~E. Bolch, C.~G. Graff, C.~H. Kim, N.~Kuster, B.~Lloyd,
  T.~Morrison, P.~Segars, Y.~S. Yeom, {\em et~al.}, ``Advances in computational
  human phantoms and their applications in biomedical engineering—a topical
  review,'' {\em IEEE transactions on radiation and plasma medical sciences},
  vol.~3, no.~1, pp.~1--23, 2018.

\bibitem{johnson2019mimic}
A.~E. Johnson, T.~J. Pollard, S.~J. Berkowitz, N.~R. Greenbaum, M.~P. Lungren,
  C.-y. Deng, R.~G. Mark, and S.~Horng, ``Mimic-cxr, a de-identified publicly
  available database of chest radiographs with free-text reports,'' {\em
  Scientific data}, vol.~6, no.~1, pp.~1--8, 2019.

\bibitem{midrc}
``Medical imaging and data resource center (midrc).''
  \url{https://www.midrc.org/}.
\newblock Accessed: 2023-01-10.

\bibitem{Sturgeon2017}
G.~M. Sturgeon, S.~Park, W.~P. Segars, and J.~Y. Lo, ``{{S}ynthetic breast
  phantoms from patient based eigenbreasts},'' {\em Med Phys}, vol.~44, no.~12,
  p.~6270–6279, 2017.

\bibitem{lewis2000pose}
J.~P. Lewis, M.~Cordner, and N.~Fong, ``Pose space deformation: a unified
  approach to shape interpolation and skeleton-driven deformation,'' in {\em
  Proceedings of the 27th annual conference on Computer graphics and
  interactive techniques}, pp.~165--172, 2000.

\bibitem{chen2020generating}
J.~Chen, Y.~Li, Y.~Du, and E.~C. Frey, ``Generating anthropomorphic phantoms
  using fully unsupervised deformable image registration with convolutional
  neural networks,'' {\em Medical physics}, vol.~47, no.~12, pp.~6366--6380,
  2020.

\bibitem{galbusera2018exploring}
F.~Galbusera, F.~Niemeyer, M.~Seyfried, T.~Bassani, G.~Casaroli, A.~Kienle, and
  H.-J. Wilke, ``Exploring the potential of generative adversarial networks for
  synthesizing radiological images of the spine to be used in in silico
  trials,'' {\em Frontiers in bioengineering and biotechnology}, vol.~6, p.~53,
  2018.

\bibitem{brock2018large}
A.~Brock, J.~Donahue, and K.~Simonyan, ``Large scale gan training for high
  fidelity natural image synthesis,'' {\em arXiv preprint arXiv:1809.11096},
  2018.

\bibitem{zhu2017unpaired}
J.-Y. Zhu, T.~Park, P.~Isola, and A.~A. Efros, ``Unpaired image-to-image
  translation using cycle-consistent adversarial networks,'' in {\em
  Proceedings of the IEEE international conference on computer vision},
  pp.~2223--2232, 2017.

\bibitem{isola2017image}
P.~Isola, J.-Y. Zhu, T.~Zhou, and A.~A. Efros, ``Image-to-image translation
  with conditional adversarial networks,'' in {\em Proceedings of the IEEE
  conference on computer vision and pattern recognition}, pp.~1125--1134, 2017.

\bibitem{wang2021generative}
Z.~Wang, Q.~She, and T.~E. Ward, ``Generative adversarial networks in computer
  vision: A survey and taxonomy,'' {\em ACM Computing Surveys (CSUR)}, vol.~54,
  no.~2, pp.~1--38, 2021.

\bibitem{goodfellow2020generative}
I.~Goodfellow, J.~Pouget-Abadie, M.~Mirza, B.~Xu, D.~Warde-Farley, S.~Ozair,
  A.~Courville, and Y.~Bengio, ``Generative adversarial networks,'' {\em
  Communications of the ACM}, vol.~63, no.~11, pp.~139--144, 2020.

\bibitem{singh2021medical}
N.~K. Singh and K.~Raza, ``Medical image generation using generative
  adversarial networks: A review,'' {\em Health informatics: A computational
  perspective in healthcare}, pp.~77--96, 2021.

\bibitem{bojanowski2017optimizing}
P.~Bojanowski, A.~Joulin, D.~Lopez-Paz, and A.~Szlam, ``Optimizing the latent
  space of generative networks,'' {\em arXiv preprint arXiv:1707.05776}, 2017.

\bibitem{ho2020denoising}
J.~Ho, A.~Jain, and P.~Abbeel, ``Denoising diffusion probabilistic models,''
  {\em Advances in Neural Information Processing Systems}, vol.~33,
  pp.~6840--6851, 2020.

\bibitem{li2020federated}
D.~Li, A.~Kar, N.~Ravikumar, A.~F. Frangi, and S.~Fidler, ``Federated
  simulation for medical imaging,'' in {\em International Conference on Medical
  Image Computing and Computer-Assisted Intervention}, pp.~159--168, Springer,
  2020.

\bibitem{croitoru2022diffusion}
F.-A. Croitoru, V.~Hondru, R.~T. Ionescu, and M.~Shah, ``Diffusion models in
  vision: A survey,'' {\em arXiv preprint arXiv:2209.04747}, 2022.

\bibitem{dhariwal2021diffusion}
P.~Dhariwal and A.~Nichol, ``Diffusion models beat gans on image synthesis,''
  {\em Advances in Neural Information Processing Systems}, vol.~34,
  pp.~8780--8794, 2021.

\bibitem{zhou2022learning}
W.~Zhou, S.~Bhadra, F.~J. Brooks, H.~Li, and M.~A. Anastasio, ``Learning
  stochastic object models from medical imaging measurements by use of advanced
  ambient generative adversarial networks,'' {\em Journal of Medical Imaging},
  vol.~9, no.~1, p.~015503, 2022.

\bibitem{graff2016new}
C.~G. Graff, ``A new, open-source, multi-modality digital breast phantom,'' in
  {\em Medical Imaging 2016: Physics of Medical Imaging}, vol.~9783,
  pp.~72--81, SPIE, 2016.

\bibitem{Bakic2016VirtualToolsfortheEvaluationofBreastImaging-State-Of-TheScienceandFutureDirections}
P.~R. Bakic, K.~J. Myers, S.~J. Glick, and A.~D. Maidment, ``Virtual tools for
  the evaluation of breast imaging: state-of-the science and future
  directions,'' pp.~518--524, Springer, Cham, 6 2016.

\bibitem{dukov2019models}
N.~Dukov, K.~Bliznakova, F.~Feradov, I.~Buliev, H.~Bosmans, G.~Mettivier,
  P.~Russo, L.~Cockmartin, and Z.~Bliznakov, ``Models of breast lesions based
  on three-dimensional x-ray breast images,'' {\em Physica Medica}, vol.~57,
  pp.~80--87, 2019.

\bibitem{bliznakova2019development}
K.~Bliznakova, N.~Dukov, F.~Feradov, G.~Gospodinova, Z.~Bliznakov, P.~Russo,
  G.~Mettivier, H.~Bosmans, L.~Cockmartin, A.~Sarno, {\em et~al.},
  ``Development of breast lesions models database,'' {\em Physica Medica},
  vol.~64, pp.~293--303, 2019.

\bibitem{Kadia}
D.~Kadia, T.~V. Nguyen, and V.~Asari, ``Synthesis for robust segmentation of
  infected lung region on small-scale data,'' {\em SSRM}, 2022.

\bibitem{de2015computational}
L.~de~Sisternes, J.~G. Brankov, A.~M. Zysk, R.~A. Schmidt, R.~M. Nishikawa, and
  M.~N. Wernick, ``A computational model to generate simulated
  three-dimensional breast masses,'' {\em Medical physics}, vol.~42, no.~2,
  pp.~1098--1118, 2015.

\bibitem{sengupta2021computational}
A.~Sengupta, D.~Sharma, and A.~Badano, ``Computational model of tumor growth
  for in silico trials,'' in {\em Medical Imaging 2021: Physics of Medical
  Imaging}, vol.~11595, pp.~1262--1270, SPIE, 2021.

\bibitem{Sengupta2021ComputationalMO}
A.~Sengupta, D.~Sharma, and A.~Badano, ``Computational model of tumor growth
  for in silico trials,'' 2021.

\bibitem{Wolberg1988GeometricTT}
G.~Wolberg, ``Geometric transformation techniques for digital images: A
  survey,'' 1988.

\bibitem{Chlap2021-xi}
P.~Chlap, H.~Min, N.~Vandenberg, J.~Dowling, L.~Holloway, and A.~Haworth, ``A
  review of medical image data augmentation techniques for deep learning
  applications,'' {\em J. Med. Imaging Radiat. Oncol.}, vol.~65, pp.~545--563,
  Aug. 2021.

\bibitem{Hesse2021-zj}
L.~S. Hesse, G.~Kuling, M.~Veta, and A.~L. Martel, ``Intensity augmentation to
  improve generalizability of breast segmentation across different {MRI} scan
  protocols,'' {\em IEEE Trans. Biomed. Eng.}, vol.~68, pp.~759--770, Mar.
  2021.

\bibitem{NEURIPS2021_7230b2b0}
K.~Tian, C.~Lin, S.~N. Lim, W.~Ouyang, P.~Dokania, and P.~Torr, ``A continuous
  mapping for augmentation design,'' in {\em Advances in Neural Information
  Processing Systems} (M.~Ranzato, A.~Beygelzimer, Y.~Dauphin, P.~Liang, and
  J.~W. Vaughan, eds.), vol.~34, pp.~13732--13743, Curran Associates, Inc.,
  2021.

\bibitem{Noh2017RegularizingDN}
H.~Noh, T.~You, J.~Mun, and B.~Han, ``Regularizing deep neural networks by
  noise: Its interpretation and optimization,'' {\em ArXiv},
  vol.~abs/1710.05179, 2017.

\bibitem{Moreno-Barea2018}
F.~J. Moreno-Barea, F.~Strazzera, J.~M. Jerez, D.~Urda, and L.~Franco,
  ``Forward noise adjustment scheme for data augmentation,'' in {\em 2018 IEEE
  Symposium Series on Computational Intelligence (SSCI)}, pp.~728--734, 2018.

\bibitem{Bae2018-kv}
H.-J. Bae, C.-W. Kim, N.~Kim, B.~Park, N.~Kim, J.~B. Seo, and S.~M. Lee, ``A
  perlin noise-based augmentation strategy for deep learning with small data
  samples of {HRCT} images,'' {\em Sci. Rep.}, vol.~8, p.~17687, Dec. 2018.

\bibitem{Omigbodun2019}
A.~O. Omigbodun, F.~Noo, M.~McNitt-Gray, W.~Hsu, and S.~S. Hsieh, ``{{T}he
  effects of physics-based data augmentation on the generalizability of deep
  neural networks: {D}emonstration on nodule false-positive reduction},'' {\em
  Med Phys}, vol.~46, pp.~4563--4574, Oct 2019.

\bibitem{pmlr-v139-fabian21a}
Z.~Fabian, R.~Heckel, and M.~Soltanolkotabi, ``Data augmentation for deep
  learning based accelerated mri reconstruction with limited data,'' in {\em
  Proceedings of the 38th International Conference on Machine Learning}
  (M.~Meila and T.~Zhang, eds.), vol.~139 of {\em Proceedings of Machine
  Learning Research}, pp.~3057--3067, PMLR, 18--24 Jul 2021.

\bibitem{Abadi2019}
E.~Abadi, W.~P. Segars, G.~M. Sturgeon, B.~Harrawood, A.~Kapadia, and E.~Samei,
  ``Modeling “textured” bones in virtual human phantoms,'' {\em IEEE
  Transactions on Radiation and Plasma Medical Sciences}, vol.~3, no.~1,
  pp.~47--53, 2019.

\bibitem{Pezeshk2015}
A.~Pezeshk, B.~Sahiner, R.~Zeng, A.~Wunderlich, W.~Chen, and N.~Petrick,
  ``Seamless insertion of pulmonary nodules in chest ct images,'' {\em IEEE
  Transactions on Biomedical Engineering}, vol.~62, no.~12, pp.~2812--2827,
  2015.

\bibitem{Pezeshk2017}
A.~Pezeshk, N.~Petrick, W.~Chen, and B.~Sahiner, ``Seamless lesion insertion
  for data augmentation in cad training,'' {\em IEEE Transactions on Medical
  Imaging}, vol.~36, no.~4, pp.~1005--1015, 2017.

\bibitem{Ghanian_2018}
Z.~Ghanian, A.~Pezeshk, N.~Petrick, and B.~Sahiner, ``Computational insertion
  of microcalcification clusters on mammograms: reader differentiation from
  native clusters and computer-aided detection comparison,'' {\em Journal of
  Medical Imaging}, vol.~5, p.~1, nov 2018.

\bibitem{gong2006computer}
X.~Gong, S.~J. Glick, B.~Liu, A.~A. Vedula, and S.~Thacker, ``A computer
  simulation study comparing lesion detection accuracy with digital
  mammography, breast tomosynthesis, and cone-beam ct breast imaging,'' {\em
  Medical physics}, vol.~33, no.~4, pp.~1041--1052, 2006.

\bibitem{Barufaldi18}
B.~Barufaldi, D.~Higginbotham, P.~R. Bakic, and A.~D.~A. Maidment, ``{OpenVCT:
  a GPU-accelerated virtual clinical trial pipeline for mammography and digital
  breast tomosynthesis},'' in {\em Medical Imaging 2018: Physics of Medical
  Imaging} (J.~Y. Lo, T.~G. Schmidt, and G.-H. Chen, eds.), vol.~10573,
  p.~1057358, International Society for Optics and Photonics, SPIE, 2018.

\end{thebibliography}
\clearpage
\end{document}